\documentclass[10pt,twocolumn,letterpaper]{article}

\usepackage[pagenumbers]{wacv} 

\definecolor{wacvblue}{rgb}{0.21,0.49,0.74}
\usepackage[pagebackref,breaklinks,colorlinks,allcolors=wacvblue]{hyperref}
\usepackage{booktabs,multirow,makecell,xcolor}
\usepackage{parskip}

\def\wacvPaperID{1528} 
\def\confName{WACV}
\def\confYear{2026}

\title{Anatomy-VLM: A Fine-grained Vision-Language Model for Medical Interpretation}
\author{Difei Gu\textsuperscript{1}, Yunhe Gao\textsuperscript{1,2}, Mu Zhou\textsuperscript{1}, Dimitris Metaxas\textsuperscript{1}\\
\textsuperscript{1}Rutgers University\\
\textsuperscript{2}Stanford University\\
}

\begin{document}
\maketitle

\begin{abstract}
Accurate disease interpretation from radiology remains challenging due to imaging heterogeneity. Achieving expert-level diagnostic decisions requires integration of subtle image features with clinical knowledge. Yet major vision-language models (VLMs) treat images as holistic entities and overlook fine-grained image details that are vital for disease diagnosis. Clinicians analyze images by utilizing their prior medical knowledge and identify anatomical structures as important region of interests (ROIs). Inspired from this human-centric workflow, we introduce Anatomy-VLM, a fine-grained, vision-language model that incorporates multi-scale information. First, we design a model encoder to localize key anatomical features from entire medical images. Second, these regions are enriched with structured knowledge for contextually-aware interpretation. Finally, the model encoder aligns multi-scale medical information to generate clinically-interpretable disease prediction. Anatomy-VLM achieves outstanding performance on both in- and out-of-distribution datasets. We also validate the performance of Anatomy-VLM on downstream image segmentation tasks, suggesting that its fine-grained alignment captures anatomical and pathology-related knowledge. Furthermore, the Anatomy-VLM's encoder facilitates zero-shot anatomy-wise interpretation, providing its strong expert-level clinical interpretation capabilities.



\end{abstract}    
\section{Introduction}
\label{sec:intro}

\begin{figure}[t]              
  \includegraphics[width=\columnwidth]{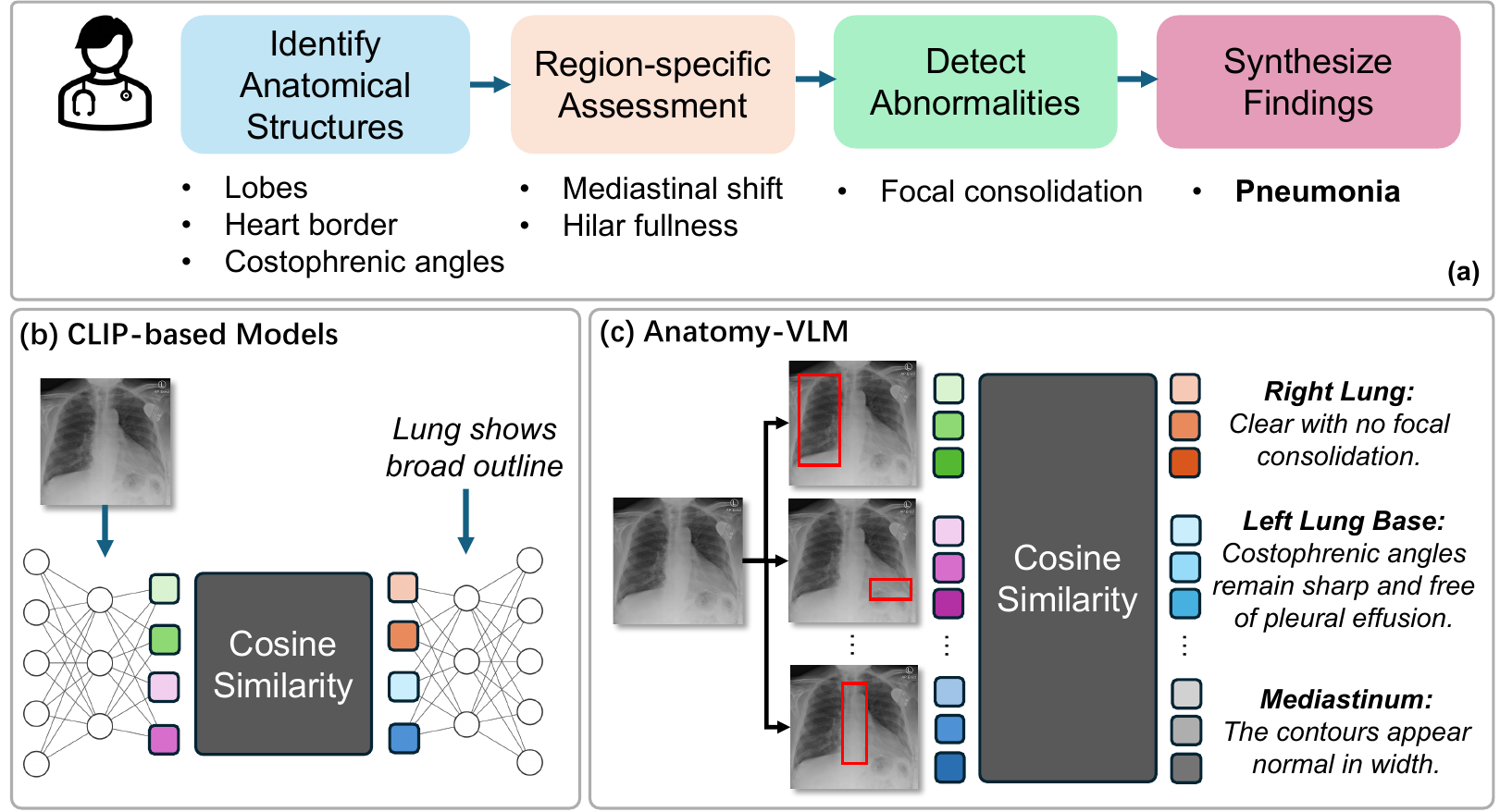}

  \caption{Overview of methodology differences among human-expert workflow, standard VLM and our Anatomy-VLM that is inspired from a radiologist's workflow. (a) Human-expert radiologist's approach first identifies anatomical structures, then performs a region-specific assessment of each area, identifies abnormalities, and finally synthesizes the findings into a coherent clinical report. (b) Conventional vision–language methods embed the entire chest X-ray as a single global feature and match it to textual concepts via cosine similarity, showing limits on spatial precision and interpretability. (c) Anatomy-VLM emphasizes the anatomy-wise contrastive learning. We partition the radiograph into clinically meaningful regions, generate a dedicated embedding for each region, and contrast these embeddings against structured anatomical concepts. Our approach delivers fine-grained, interpretable predictions that align with the radiologist’s interpretation pipeline.}
  \label{fig:compare}
\end{figure}

Medical image interpretation is crucial for disease diagnosis, monitoring, and treatment planning \cite{zhao2023clip}. Current efforts based on vision-language pre-training (VLP) \cite{zhang2024data,zhang2024vision,radford2021learning,boecking2022making,gao2024aligning} demonstrate promising performance in image and text alignment, achieving impressive zero-shot accuracy in whole-image to text matching. A typical approach is to match the whole-image input to the associated text descriptions via contrastive learning \cite{radford2021learning}. While these methods excel at surface-level data alignment, it remains unclear whether the encoder can truly capture and express the understanding of disease relations. In real-world clinics, professional radiologists do not make decisions solely on image sequences; instead, they often engage in a systematic multi-step reasoning process \cite{waite2019analysis,gu2025radalign}. In Fig.~\ref{fig:compare} (a), this multi-step operation involves identifying anatomical structures, recognizing their common attributes, and utilizing prior medical knowledge to gain a contextual understanding. This anatomical interpretation complexity is exemplified in chest X-ray pneumonia (lung inflammation) diagnosis \cite{walker2014imaging}, where radiologists must first identify precise anatomical boundaries of each lung lobe, then assess the characteristic air-filled appearance of healthy lung tissue versus areas of consolidation. This process leverages the inherent value of human-expert workflow, following the anatomical structure localization, regional correlation, and comprehensive medical knowledge integration to generate clinically-validated and explainable findings.

Recent approaches \cite{wu2023medklip,lai2024carzero} have advanced language processing to enable rich semantic representation and image interpretation. Yet these methods fundamentally adopt a global image-to-text alignment, overlooking the anatomically-grounded nature of disease assessment. First, the global-level alignment creates semantic ambiguity especially when identical medical concepts are jointly applied to multiple anatomical regions. For instance, the term ``consolidation" in radiology reports may refer to lung consolidation in the upper lobe or lower lobe, each carrying distinct clinical implications and treatment considerations. Second, the absence of anatomy-wise alignment likely leads to knowledge conflation, where pathological findings from different anatomical structures become incorrectly associated. For instance, cardiomegaly (enlarged heart) and pulmonary edema (fluid in lungs) frequently co-occur in chest X-rays, but represent distinct pathophysiological processes affecting different organs. Therefore, a global alignment can spuriously correlate textual descriptions of cardiac abnormalities with pulmonary regions, leading to a diagnostic confusion and erroneous interpretation. 

\begin{figure*}[t]              
  \centering
  \includegraphics[width=0.9\textwidth]{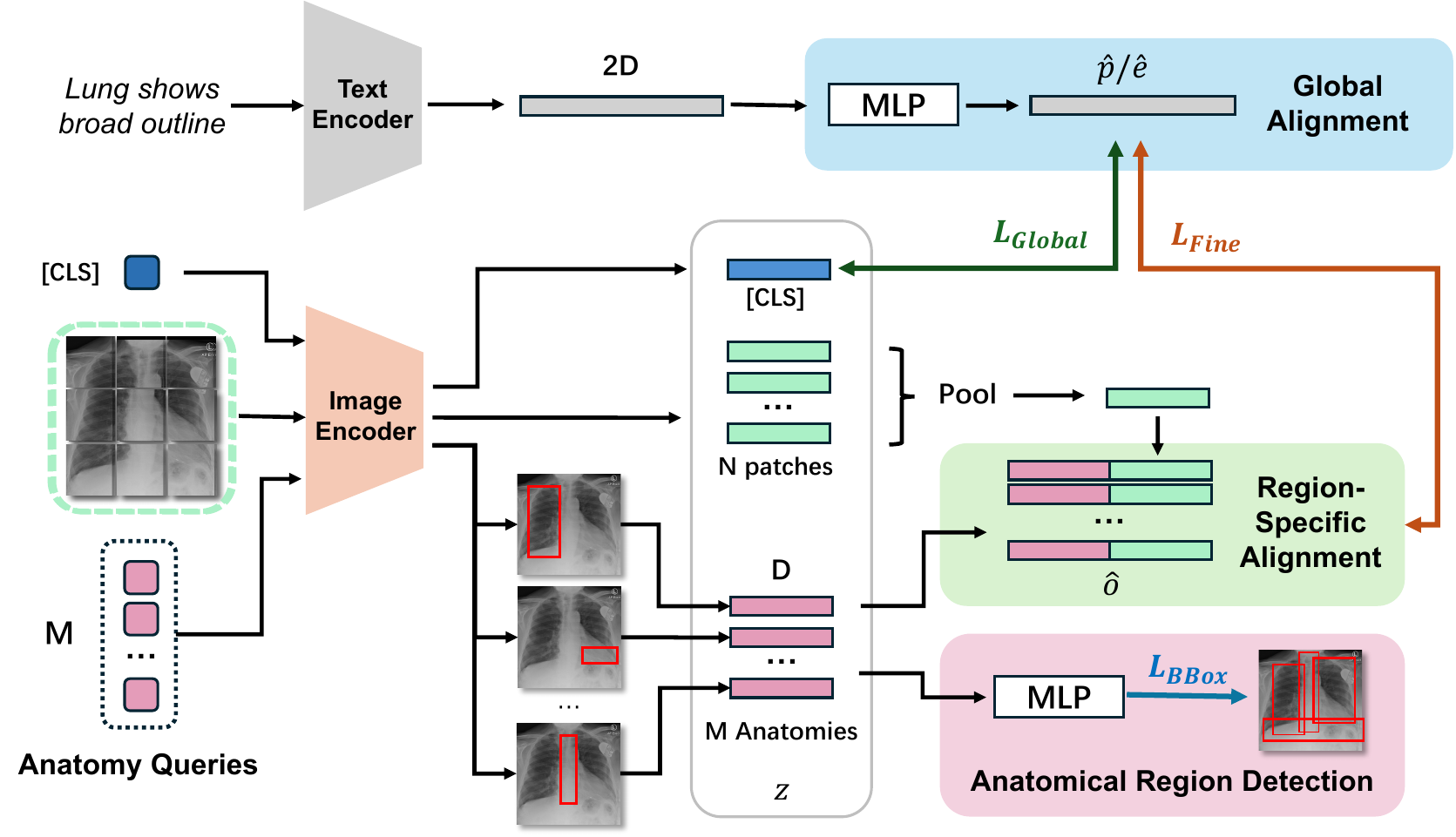}

  \caption{Overview of Anatomy-VLM. The model processes medical images and clinical text to generate global and region-specific alignments, using anatomy queries to detect and localize anatomical structures. We design the model to follow human expert diagnostic interpretation with three components: \textbf{Anatomical Region Detection}, \textbf{Region-specific Alignment}, and \textbf{Global Alignment}.}
  \label{fig:main-pipeline}
\end{figure*}


To address the above hurdles and enable anatomically-aware representation learning, we present in this paper \textbf{Anatomy-VLM}, a novel medical visual-language model (VLM) enhancement that integrates a critical procedure that commonly arises in human-expert workflows. Specifically, Anatomy-VLM extracts region-specific queries from radiology images using an object detection encoder, enabling the model to localize anatomical structures and extract pathological features. To enhance outcome interpretation, we develop structured medical knowledge integration to enrich image content as illustrated in Fig. \ref{fig:compare} (c). Overall, the anatomy-knowledge alignment emphasizes fine-grained clinical knowledge integration and understanding. The key contributions of our work are as follows:

\begin{itemize}
    \item We introduce a novel VLM approach that extracts region-based image features and enables fine-grained anatomical and pathological localization.
    \item We propose an expert-aligned training pipeline that correlates with radiologist's workflows, demonstrating that the model can effectively understand complex disease relationships through structured anatomical interpretation.
    \item We provide extensive validation on fine-grained zero-shot disease detection datasets and downstream segmentation tasks, showing that Anatomy-VLM improves accuracy and robustness across long-tail data distributions.
\end{itemize}
\section{Related Works}
\label{sec:related_works}


\subsection{Vision-Language Pre-training}

Vision-Language Models (VLMs) \cite{radford2021learning,li2023blip,jia2021scaling,chen2025cost,gao2025show,chen2024survey} have emerged as a powerful paradigm for aligning visual and textual representation through multimodal pre-training. While general-purpose VLMs such as CLIP \cite{radford2021learning}, BLIP \cite{li2023blip}, and LLaVA \cite{liu2023visual} demonstrate impressive performance on natural images, they often lack the specialized medical knowledge for disease understanding and clinical interpretation. To address this hurdle, domain-specific model adaptations become increasingly useful. For instance, BioViL \cite{boecking2022making} leverages radiology image-text pairs for vision-language alignment. MedCLIP \cite{wang2022medclip} focuses on medical image-text contrastive learning with clinical report supervision. MedKLIP \cite{wu2023medklip} incorporates a medical knowledge-enhanced triplet to improve feature representation learning. Although these VLMs are good at learning multimodal representation, they are based mainly on the well-designed image-text matching during pre-training. These approaches predominantly treat medical images holistically in the data alignment process, lacking explicit modeling of the detectable anatomical structures. In clinics, radiologists must rely on the fine-grained anatomical localization and knowledge integration \cite{datta2020understanding,waite2019analysis} to inform decision making. Inspired by this human-expert workflow, Anatomy-VLM addresses the fundamental gap between image feature extraction and human workflows by incorporating anatomical interpretation with structured knowledge integration. Unlike prior end-to-end classification \cite{dosovitskiy2020image, boecking2022making}, Anatomy-VLM closely mimics the multi-step reasoning process that radiologists employ in clinical practice. By modeling the regional visual-and-text relationship, Anatomy-VLM achieves strong predictions that address the challenge of long‑tail class imbalance and generalize well to rare conditions.

\subsection{Fine-grained VLM}
Fine-grained alignment in medical vision-language models has gained attention to address the limitation of global image-text matching \cite{lu2025integrating}. For instance, fine-grained vision-language models (fVLM) \cite{shui2025large} matches anatomical regions of 3D CT images with corresponding descriptions. MedKLIP \cite{wu2023medklip} leverages an entity-level alignment with corresponding image regions. CARZero \cite{lai2024carzero} introduces an alignment that converts cross‑attention features into similarity representations. RGRG \cite{tanida2023interactive} achieves fine-grained alignment by dividing images into a fixed grid of predefined regions for region-specific visual-textual matching. ASG \cite{li2024anatomical} leverages anatomical structure guidance for vision-language pre-training. However, these approaches primarily rely on implicit alignment strategies through contrastive learning, focusing on either image-text matching or region-text correspondence in isolation.
By contrast, Anatomy-VLM introduces a fundamentally different alignment paradigm that unifies object locations, visual features, and textual information through concept queries. This trimodal integration enables the model to simultaneously learn spatial-semantic correspondences, where object locations inform visual attention, visual features ground textual descriptions, and textual semantics guide spatial localization—creating a synergistic learning process that localizes clinically meaningful anatomical structures, providing precise region identification. Furthermore, Anatomy-VLM incorporates multi-scale information processing that mirrors the clinical diagnostic workflow, seamlessly integrating information from fine-grained localization of anatomical structures to global pathological classification. This hierarchical approach ensures that local anatomical findings inform broader diagnostic decisions, while global context guides the interpretation of localized features. As a result, Anatomy-VLM aims to deliver clinically-grounded outputs with transparent interpretability across multiple scales of medical reasoning.


\section{Method}

\subsection{Problem Formulation}
Clinical image assessment essentially requires both anatomical structure and domain-specific knowledge to reach reliable diagnostic conclusions. We formulate this as a granular medical vision-language alignment problem that unifies anatomical region detection, knowledge-enhanced alignment, and disease-level classification. We consider a medical image \(\mathcal{I}\in \mathbb{R}^{H\times W\times C}\) from any modality. We introduce a set of \(M\) learnable anatomical bounding-box queries \(\mathcal{A} \in \{a^k\}^M_{k=1}\) that serve as spatial anchors that governs \(M\) clinically relevant anatomical structures. These anatomical anchors are designed to capture the systematic spatial features that medical professionals identify when analyzing medical images. For each anatomical region, we associate it with structured clinical descriptions. We collect all phrases  \(\mathcal{P}\in \{p^k\}^m_{k=1}\) or disease categories \(\mathcal{D}\in \{d^k\}^n_{k=1}\) for image \(\mathcal{I}\) where every phrase \(p^j\) or class label \(d^j\) describes anatomical findings within a specific anatomical region \(a^j\) to incorporate anatomy level expertise. Each medical image is associated with a multi-label disease vector \(\mathcal{Y} \in \{0, 1\}^C\) where \(C\) represents the number of possible disease categories. Our objective is to learn an encoder \(f_\theta\) that performs anatomy-aware diagnostic interpretation by mapping the input tuple \((\mathcal{I}, \mathcal{P},  \mathcal{A})\) to refined anatomical predictions \((\hat{a}^k, \hat{b}^k, \hat{p}^k)\) and final disease prediction \(\hat{y}\), where \(\hat{a}^k\) represents anatomical queries, \(\hat{b}^k\) represents knowledge-enhanced regional observations, and \(\hat{p}^k\) is the aligned clinical phrase embedding.

\subsection{Anatomical Region Detection} 
In Fig.~\ref{fig:main-pipeline}, Anatomy-VLM utilizes a vision transformer backbone that processes a visual embedding \(z\) that contains [CLS] token, \(L\) patch tokens, and  \(M\) anatomical tokens through a self-attention mechanism. The anatomy queries serve as learnable tokens that enable the model to automatically detect and localize clinically-relevant anatomical structures in chest X-ray images. Each anatomical query \(a^k \in \mathbb{R}^d\) is designed to encode region-specific clinical evidence corresponding to its designed anatomical region. 

The anatomical detection process is supervised through a set-prediction formulation that combines spatial localization and anatomical structure classification. The anatomical detection loss \(\mathcal{L}_{anat}\) employs a combination of GIOU and L1 loss to ensure accurate bounding box regression:
\begin{equation}
\mathcal{L}_{anat}
= \frac{1}{M}\sum_{k=1}^{M} [\text{GIoU}(\hat{b}^k_i, b^{k}_i) + \alpha \text{L1}(\hat{b}^k_i, b^{k}_i)],
\end{equation}
where \(\hat{b}^k_i\) and \(b^k_i\) represent the predicted and target anatomical bounding boxes respectively and \(\alpha\) weights the two losses. This objective ensures that each anatomical query learns to accurately localize its corresponding anatomical regions while maintaining spatial precision across different anatomical structures.

\subsection{Region-specific Alignment}
Anatomy-VLM implements an anatomy-level, region-specific alignment. A medical pre-trained text encoder \(h_\phi\) is employed to bridge the gap between visual observations and clinical language. The text encoder maps clinical descriptions \(p\) or disease category \(d\) to an embedding \((\hat{p}, \hat{e}) \in \mathbb{R}^d\) in the same representation space as the visual embedding \(z\). The learned anatomical representation \(z\) is aligned with anatomy-level phrase embeddings \(\hat{p}\) through contrastive learning, ensuring that each region establishes correspondence with clinical descriptions. To enhance the capacity of regions with global image context, we employ a global pooling and concatenation strategy. Specifically, after the patch tokens and bounding box tokens pass through linear projection layers, the \(N\) patch tokens are average-pooled from their original dimension to a global representation. This global patch representation is then concatenated with each of the bounding box tokens along the feature dimension, resulting in enriched anatomical representations \(\hat{o}\). This global pooling and concatenation enable the model to capture the localized anatomical features (i.e., encoded by individual bounding box tokens) and the broader image context provided by the average patch features. 

The anatomy-level matching loss \(\mathcal{L}_{fine}\) establishes fine-grained alignment between anatomical regions and their corresponding clinical descriptions through contrastive learning. We adopt an InfoNCE objective over the joint batch for a strong learning scope:
\begin{equation}
\mathcal{L}_{fine}
= -\frac{1}{|\mathcal{B}|}\sum_{i=1}^{B}\frac{1}{T_i} \sum_{j=1}^{T_i}\text{H}(y^{fine}_{i,j}, \langle \hat{o}_{i,j}, \hat{p}_{i,j} \rangle ),
\end{equation}
where \(T\) is the number of anatomy bounding box anchors, \(\mathcal{B}\) is the batch size, and H is the cross-entropy loss. \(y^{fine}\) denotes the ground-truth similarity label, and section \ref{sec:label_generation} discusses in more details on how it is constructed. \(\hat{p}_{i,j}\) is the contrastive similarity score computed between anatomical representations and clinical phrase embeddings. This objective drives each enriched anatomical representation \(\hat{o}\) to align closely with its corresponding knowledge-enhanced clinical phrase embedding \(\hat{p}\) while maintaining separation from unrelated descriptions, enabling precise spatial-textual correspondence at the anatomical level.

\subsection{Anatomy-Level Similarity Label Generation}
\label{sec:label_generation}
To enable effective contrastive learning from bounding box findings, it is necessary to ensure a structured representation of positive and negative trainable examples. In our study, we carefully transform raw textual findings into the structured positive and negative pairs. By design, we focus on findings from 29 bounding boxes, generating contrastive pairs \((T_i, R_i, l_i)\) where \(T_i\) represents the bounding box, \(R_i\) indicates the processed text, and \(l_i\) stores the contrastive label.
The first phase splits each finding into sub-sentences and creates initial contrastive labels. For non-empty findings, we randomly select one sub-sentence and assign a positive label \(l_i = 1\), while empty findings receive negative labels \(l_i = 0\). The second phase applies controlled perturbations to positive samples with 20\% probability, by negating expressions to improve model robustness while preserving data integrity in 80\% of cases. In the third phase, to maximize the label coverage, we address bounding boxes with no findings by generating attribute-based negative examples. To do so, we identify available attributes from positive samples and create either semantically negative content or hard negatives (positive content with negative labels) for the comprehensive dataset coverage. The final phase resolves text duplicates by augmenting the labels accordingly. This controlled augmentation maintains semantic coherence between visual regions and textual descriptions for conflict-free knowledge contrastive learning.

\subsection{Image-level Disease Classification}
To detect abnormalities for disease classification, our approach establishes correspondence between visual representations and clinical descriptions at two complementary levels, the Anatomy-level and the Image-level. In the Vision Transformer, the [CLS] token \(g \in \mathbb{R}^d\) captures holistic diagnostic information that synthesizes findings across all anatomical regions, enabling image-level clinical interpretation. For disease classification, we leverage the global [CLS] token to predict image-level pathological conditions through multi-label classification. Contrastive learning is used where the rows are disease class embedding \(\hat{e}\) and columns are [CLS] tokens \(g\), facilitating disease-level prediction through learned medical concept associations. The disease-level classification loss \(\mathcal{L}_{\text{global}}\) is computed using sigmoid cross-entropy between the similarity scores of the [CLS] embedding and disease label embeddings:
\begin{equation}
\begin{split}
\mathcal{L}_{\text{global}} = -\frac{1}{C}\sum^C_{c=1}[y^{global}_{i,c}\text{log}\sigma(\langle g_i, \hat{e}_i\rangle) + \\(1-y^{global}_{i,c})\text{log}(1-\sigma(\langle g_i, \hat{e}_i\rangle))],
\end{split}
\end{equation}
where \(\sigma\) is the sigmoid function, \(y^{global}\) is the disease class ground truth, \(g_i\) is the [CLS] token representation, and \(\hat{e}_i\) is the text embedding for disease category \(d\).

\subsection{Training Objectives}
Anatomy-VLM is trained through a multi-task objective that combines anatomical region detection, region-specific matching, and image-level disease classification. The overall loss function is formulated as:
\begin{equation}
\mathcal{L}
= \lambda_{anat}\mathcal{L}_{anat} + \lambda_{fine}\mathcal{L}_{fine} + \lambda_{global}\mathcal{L}_{global},
\end{equation}
where the weighting parameters \(\lambda\) balance the contribution of each component during training. Adjusting these weights allows for fine-tuning the model's emphasis on anatomical localization accuracy, precision of anatomical structures to pathological findings, and the model's overall disease classification performance through image-level representation learning.

These formulations enable explicit modeling of the clinical diagnostic process through anatomically-grounded interpretation with structured medical knowledge integration, addressing the fundamental limitations of existing holistic VLMs that fail to capture region-specific disease characteristics from human-expert medical diagnosis.
\section{Experiments}

\subsection{Experimental Setup}
We first evaluate Anatomy-VLM on both in-distribution and out-of-distribution zero-shot classification tasks and compare across major zero-shot and supervised models. In addition, we conduct experiments on the downstream segmentation task to evaluate the fine-grained localization capability of our pre-trained encoder. Finally, we present ablation studies and fine-grained evaluation to validate the importance of anatomy-level learning. 

\textbf{Dataset.} We train and evaluate on the public \textbf{Chest ImaGenome} \cite{wu2021chest}. Chest Imagenome builds on top of the MIMIC-CXR \cite{johnson2024mimic} dataset with over 240k frontal chest X-rays. For every image, it provides 29 standardized anatomical regions with precomputed bounding boxes and a set of finding labels extracted from the accompanying radiology report. Reports are first scanned with a rule-based NER to detect both finding and anatomy mentions (e.g., “consolidation,” “pleural effusion,” “right lower lobe”); a dependency parser then links each finding to the anatomy mentions that occur in the same sentence (i.e., “nearest” in the syntactic sense). The anatomy phrases are normalized to one of the 29 predefined regions, which are linked to image coordinates via an atlas-based bounding-box pipeline (refined with a detector). This yields per-image relations of the form \emph{anatomy has/has-no finding} that populate the dataset’s scene graph. For the external dataset, we use \textbf{IU-Xray (OpenI)} \cite{demner2016preparing}, which is unseen by all comparison models. IU Xray contains 7,470 chest X-ray images with corresponding reports from 3,955 patients. For the segmentation downstream task, we adopt segmentation datasets CheXmask \cite{gaggion2024chexmask} and SIIM-ACR \cite{siim-acr-pneumothorax-segmentation}. CheXmask based on MIMIC-CXR \cite{johnson2024mimic} provides segmentation labels on heart segmentation. SIIM-ACR, based on Chest X-ray 14, provides segmentation label for the disease Pneumonia. We aim to use a diversification of the datasets to showcase Anatomy-VLM's pre-train robustness.

\textbf{Baseline models.} We compare Anatomy-VLM with two classes of models. Contrastive pre-train encoders CLIP \cite{radford2021learning}, BiomedClip \cite{zhang2023biomedclip}, BioViL \cite{boecking2022making}, MedKLIP \cite{wu2023medklip}, and CARZero \cite{lai2024carzero}. And supervised classification models, including ResNet50 \cite{he2015deep} and ViT \cite{dosovitskiy2020image} models. All zero-shot models are used in their off-the-shelf configuration, while the black-box models are trained on the chest imagenome, validated on the IU X-ray dataset.

\textbf{Implementation details.} We train Anatomy-VLM by adding the supervision losses in stages. We train the encoder with the object detection supervision only in the first stage. In the second stage the detection specialized encoder is jointly trained with the global loss, and finally we train for all the losses in the third stage. During training, we only optimize the visual encoder, anatomy queries, and all the linear layers with AdamW optimizer for 40 epoches, using a decreasing learning rate of [1e-3, 1e-4], while keeping the text encoder fixed. For the \(\lambda\), we use the value of [1,1,1] for a balanced objective fine-tune. We implement a text-based augmentation to ensure contrastive diversity (details are provided in the supplementary). All experiments are conducted using PyTorch with Nvidia RTX 8000 GPUs.

\begin{table*}[t]
  \centering
  \caption{In-distribution zero-shot image-level disease classification performance on Chest ImaGenome. Values in percent \%.}
  \setlength{\tabcolsep}{4pt}
  \renewcommand\theadfont{\normalsize\bfseries}
  \resizebox{\textwidth}{!}{%
  \begin{tabular}{l|ccccccccccccccc|cccccc|ccc}
    \toprule
    \multirow{3}{*}{\textbf{Attribute}} & \multicolumn{15}{c}{\textbf{Vision-Language Pretrain Models}} & \multicolumn{6}{c}{\textbf{Supervised Training Models}} & \multicolumn{3}{c}{\textbf{Fine-grain Alignment}} \\[2pt]
    & \multicolumn{3}{c}{CLIP}& \multicolumn{3}{c}{BioMedCLIP}& \multicolumn{3}{c}{BioViL}& \multicolumn{3}{c}{MedKLIP}& \multicolumn{3}{c}{CARZero} & \multicolumn{3}{c}{ResNet}& \multicolumn{3}{c}{ViT-B}& \multicolumn{3}{c}{Ours} \\[-2pt]
    \cmidrule(lr){2-4}\cmidrule(lr){5-7}\cmidrule(lr){8-10}\cmidrule(lr){11-13}
    \cmidrule(lr){14-16}\cmidrule(lr){17-19}\cmidrule(lr){20-22}\cmidrule(lr){23-25}
    & BMAC & AUC & F1 & BMAC & AUC & F1 & BMAC & AUC & F1 & BMAC & AUC & F1 & BMAC & AUC & F1 & BMAC & AUC & F1 & BMAC & AUC & F1 & BMAC & AUC & F1 \\ \midrule
    Lung parenchyma \& air-space disease & 52.0 & 52.0 & 21.1 & 55.5 & 55.5 & 15.9 & 70.0 & 70.0 & 30.8 & 52.9 & 58.0 & 16.0 & 60.7 & 82.8 & 26.9 & 57.7 & 61.4 & 19.6 & 65.2 & 74.8 & 35.6 & 69.4 & 79.0 & 36.6 \\
    
    Atelectasis \& collapse & 51.1 & 51.1 & 11.5 & 53.4 & 53.4 & 8.4 & 66.0 & 66.0 & 23.7 & 50.6 & 50.6 & 24.1 & 62.9 & 83.5 & 26.2 & 59.1 & 65.0 & 32.5 & 64.3 & 77.0 & 31.5 & 67.1 & 80.0 & 32.6 \\
    
    Pleural space \& pleura & 50.1 & 50.1 & 14.1 & 50.8 & 50.8 & 4.1 & 67.7 & 67.7 & 26.6 & 51.1 & 62.1 & 16.1 & 62.2 & 85.8 & 26.2 & 62.3 & 66.6 & 24.2 & 69.1 & 78.2 & 32.4 & 69.2 & 81.7 & 35.0 \\
    
    Inflation \& airway mechanics & 50.0 & 50.0 & 10.0 & 50.1 & 50.1 & 0.6 & 58.4 & 58.4 & 12.0 & 50.0 & 43.4 & 5.9 & 78.1 & 90.6 & 45.4 & 56.8 & 56.7 & 8.2 & 71.0 & 87.4 & 39.7 & 74.3 & 89.8 & 46.3 \\
    
    Diaphragm \& sub-diaphragmatic & 49.9 & 49.9 & 0.0 & 52.7 & 52.7 & 7.5 & 63.2 & 63.2 & 7.1 & 58.7 & 65.6 & 4.6 & 77.1 & 88.3 & 40.6 & 55.7 & 63.5 & 8.0 & 72.5 & 85.8 & 35.6 & 66.1 & 84.8 & 32.9 \\
    
    Cardiomediastinal \& hilar structures & 49.4 & 49.4 & 4.7 & 51.0 & 51.0 & 6.4 & 69.1 & 69.1 & 25.7 & 50.3 & 57.6 & 14.6 & 68.6 & 88.0 & 37.8 & 61.7 & 69.5 & 24.7 & 69.4 & 81.1 & 35.5 & 69.8 & 83.4 & 36.0 \\
    
    Musculoskeletal \& thoracic cage & 50.0 & 50.0 & 2.2 & 50.1 & 50.1 & 1.0 & 52.1 & 52.1 & 6.4 & 49.6 & 44.0 & 6.0 & 69.4 & 83.0 & 24.6 & 60.3 & 65.6 & 12.7 & 61.6 & 76.0 & 22.6 & 63.9 & 76.8 & 19.2 \\
    
    \midrule
    Average & 50.5 & 50.5 & 11.0 & 52.4 & 52.4 & 8.0 & 66.1 & 66.1 & 23.3 & 51.4 & 55.7 & 14.8 & 65.8 & \textbf{85.3} & 30.8 & 59.7 & 65.0 & 21.7 & 67.0 & 78.5 & 33.4 & \textbf{68.7} & 81.2 & \textbf{34.2} \\
    \bottomrule
  \end{tabular}}
  \label{tab:imagenome_noAcc}
\end{table*}

\begin{table*}[t]
\centering
\caption{Out-of-distribution zero-shot image-level disease classification on the IU Chest X-ray (OpenI). Values in percent \%.}
\setlength{\tabcolsep}{4pt}
\resizebox{\textwidth}{!}{%
\begin{tabular}{l|ccccccccccccccc|cccccc|ccc}
\toprule
\multirow{3}{*}{\textbf{Attribute}} & \multicolumn{15}{c}{\textbf{Vision-Language Pretrain Models}} & \multicolumn{6}{c}{\textbf{Supervised Training Models}} & \multicolumn{3}{c}{\textbf{Fine-grain Alignment}} \\[2pt]
\multirow{2}{*}{Disease} &
\multicolumn{3}{c}{CLIP} &
\multicolumn{3}{c}{BioMedCLIP} &
\multicolumn{3}{c}{BioViL} &
\multicolumn{3}{c}{MedKLIP} &
\multicolumn{3}{c}{CARZero} &
\multicolumn{3}{c}{ResNet} &
\multicolumn{3}{c}{ViT} &
\multicolumn{3}{c}{Ours}\\
\cmidrule(lr){2-4}\cmidrule(lr){5-7}\cmidrule(lr){8-10}\cmidrule(lr){11-13}
\cmidrule(lr){14-16}\cmidrule(lr){17-19}\cmidrule(lr){20-22}\cmidrule(lr){23-25}
 & BMAC & AUC & F1
 & BMAC & AUC & F1
 & BMAC & AUC & F1
 & BMAC & AUC & F1
 & BMAC & AUC & F1
 & BMAC & AUC & F1
  & BMAC & AUC & F1
 & BMAC & AUC & F1\\
\midrule
Edema                & 44.3 & 44.3 &  4.0 & 50.0 & 50.0 &  4.6 & 76.9 & 87.6 & 25.0 & 50.9 & 55.0 & 24.7 & 58.7 & 94.0 & 29.5 & 67.2 & 70.5 & 38.5 & 67.9 & 75.7 & 47.3 & 78.1 & 89.7 & 60.7\\
Pleural Effusion     & 50.8 & 50.8 & 10.2 & 58.3 & 58.3 & 28.6 & 79.1 & 88.1 & 53.6 & 50.6 & 76.3 & 23.1 & 58.8 & 96.4 & 29.6 & 50.2 & 35.8 & 23.0 & 55.4 & 52.5 & 23.8 & 58.4 & 56.5 & 27.0\\
Atelectasis          & 46.1 & 46.1 & 15.7 & 51.7 & 51.7 & 15.9 & 68.9 & 68.9 & 44.4 & 52.1 & 68.0 &  2.4 & 74.6 & 98.0 & 44.4 & 62.1 & 83.7 & 25.0 & 74.6 & 93.5 & 44.4 & 93.4 & 96.9 & 36.4\\
Cardiomegaly         & 50.0 & 50.0 & 22.9 & 52.3 & 52.3 &  8.7 & 76.2 & 76.3 & 55.7 & 50.0 & 34.6 &  4.6 & 68.1 & 97.0 & 40.0 & 82.5 & 87.2 & 25.5 & 73.6 & 89.7 & 38.1 & 62.5 & 90.5 & 40.0 \\
Consolidation        & 49.6 & 49.6 &  0.0 & 49.3 & 49.3 &  0.0 & 70.1 & 70.1 & 16.7 & 50.6 & 69.8 & 10.2 & 66.7 & 97.2 & 50.0 & 81.4 & 83.3 & 30.3 & 70.2 & 77.0 & 26.0 & 79.6 & 83.3 & 35.6\\
\midrule
Average              & 48.1 & 48.1 & 10.6 & 52.3 & 52.3 & 11.6 & 74.2 & 80.0 & 39.1 & 50.8 & 60.8 & 13.0 & 65.4 & \textbf{96.5} & 38.7 & 68.7 & 72.1 & 28.5 & 68.3 & 77.7 & 35.9 & \textbf{74.4} & 83.4 & \textbf{39.9}\\
\bottomrule
\end{tabular}}%
\label{tab:iuxray_valid}
\end{table*}

\subsection{Model Comparison}
\textbf{Zero-shot classification on in-distribution data.} We evaluated Anatomy-VLM's performance on 20 clinically important fine-grained disease classes from the Chest ImaGenome dataset. We selected 20 classes rather than the 14 classes typically used from the MIMIC-CXR dataset, to demonstrate the model's ability to generalize to long-tail disease categories. Table \ref{tab:imagenome_noAcc} presents results organized by anatomical groups that map individual diseases to seven higher-level categories. Detailed long-tail analysis and the complete results can be found in the supplementary. 

Our method achieves an average BMAC of 68.7 and F1 of 34.2, significantly outperforming pretrained vision-language baselines that focus solely on aligning image-level features and texts (BioMedCLIP and MedKLIP). Anatomy-VLM's superior BMAC and F1 scores indicate better handling of class imbalance scenario, a critical challenge in medical imaging diagnosis. Although CARZero achieves the leading AUC performance, it is noted that the AUC can be inaccurate in imbalanced datasets due to the large number of true negatives. In contrast, BMAC and F1 are more sensitive to class imbalance and faithfully reflect real-world clinical performance where rare diseases must be detected accurately. In addition, we observe that Anatomy-VLM outperforms the fully supervised baselines of ResNet50 and ViT throughout the spectrum of chest abnormalities. The consistent improvement suggests that concept alignment offers a robust framework that combines the explainability benefits of vision-language models while remaining effective against imbalanced and long-tail data distributions.

\textbf{Zero-shot classification on out-of-distribution data.} We further evaluated Anatomy-VLM on the IU Chest X-ray validation set, which contains 340 studies across five essential disease classes. In Table \ref{tab:iuxray_valid}, our method demonstrates strong generalization with an average performance of 74.4 BMAC and 83.4 AUC, outperforming all zero-shot vision-language baselines and remaining competitive with supervised baselines. The consistent performance across this external dataset validates the robustness of our concept alignment approach, demonstrating that the learned representations generalize well beyond the training distribution.

\begin{table*}[t]
\centering
\caption{Segmentation performance on \textbf{CheXmask} (heart) and \textbf{SIIM-ACR} (pneumonia).
“Frozen” = encoder frozen; “Transfer” = full fine-tuning.
Best value in each row is \textbf{bold}.}
\begin{tabular*}{0.9\textwidth}{@{\hskip 6pt\extracolsep{\fill}}lllcccccc@{\hskip 6pt}}
\toprule
Dataset & Training & Metric & CLIP & BioViL & BioMedCLIP & MedKLIP & Ours \\ 
\midrule
\multirow{4}{*}{\makecell[l]{CheXmask\\Heart}}
    & \multirow{2}{*}{Frozen}   & Dice & 0.911 & 0.856 & 0.920 & 0.934 & \textbf{0.946} \\
    &                           & mIoU & 0.912 & 0.864 & 0.920 & 0.934 & \textbf{0.945} \\[2pt]
    & \multirow{2}{*}{Transfer} & Dice & 0.954 & 0.905 & 0.947 & 0.944 & \textbf{0.961} \\
    &                           & mIoU & 0.952 & 0.906 & 0.945 & 0.943 & \textbf{0.960} \\
\midrule
\multirow{4}{*}{\makecell[l]{SIIM-ACR\\Pneumonia}}
    & \multirow{2}{*}{Frozen}   & Dice & 0.000 & 0.085 & 0.051 & 0.202 & \textbf{0.243} \\
    &                           & mIoU & 0.489 & 0.487 & 0.501 & 0.541 & \textbf{0.554} \\[2pt]
    & \multirow{2}{*}{Transfer} & Dice & 0.301 & 0.174 & 0.163 & 0.232  & \textbf{0.347} \\
    &                           & mIoU & 0.577 & 0.527 & 0.532 & 0.552 & \textbf{0.603} \\
\bottomrule
\end{tabular*}
\label{tab:segmentation}
\end{table*}

\begin{figure*}[t]              
  \centering
  \includegraphics[width=0.9\textwidth]{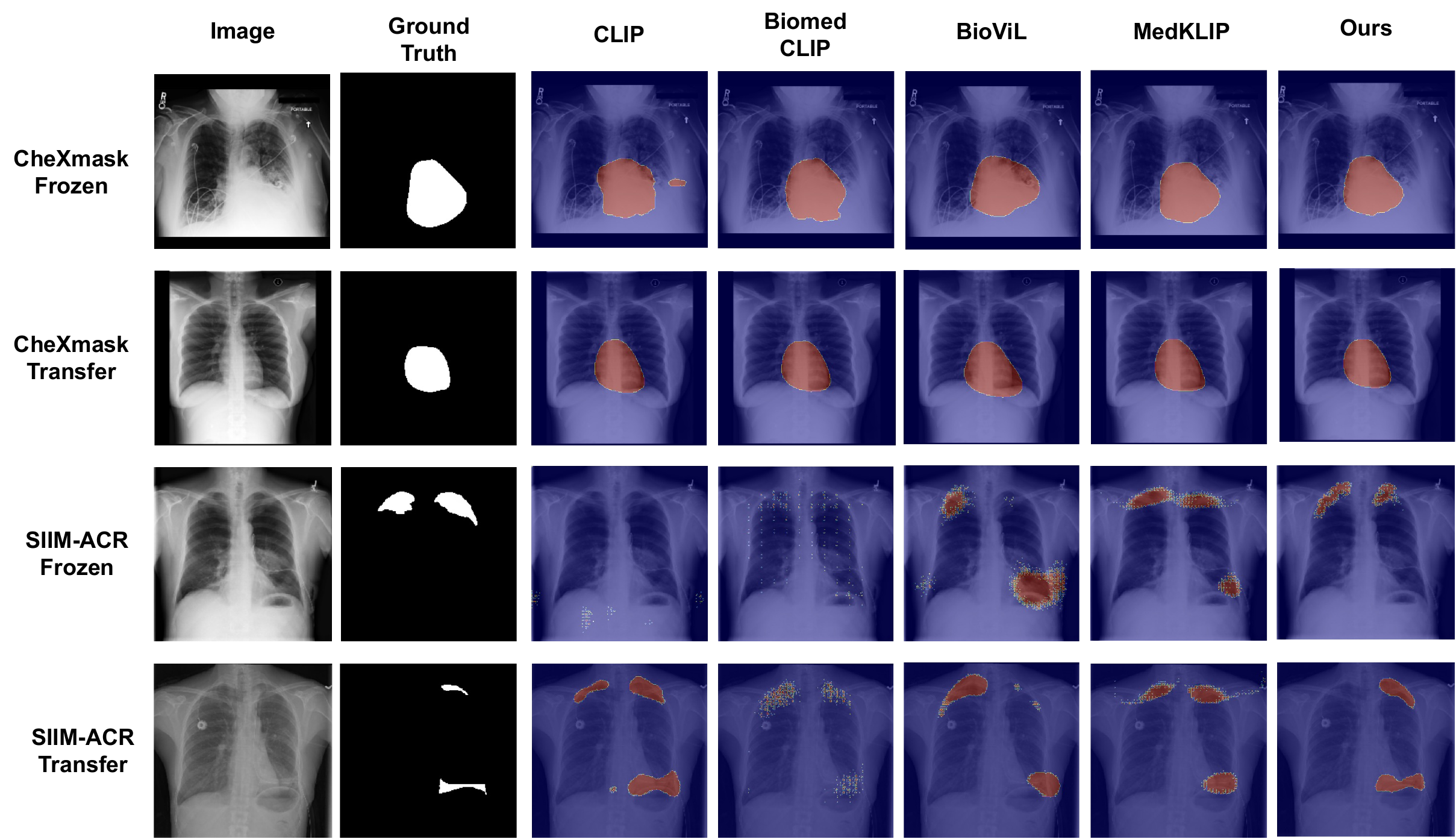}

  \caption{Evaluation of VLP encoders on segmentation tasks. The figure shows segmentation results for chest X-ray images under four different conditions: ChexMask Frozen, ChexMask Transfer, SIIM-ACR Frozen, and SIIM-ACR Transfer. For each condition, the original image and ground truth mask are shown alongside segmentation outputs from five different methods: CLIP, Biomed CLIP, BioViL, MedKLIP and Anatomy-VLM (Ours). The segmentation masks are displayed as heatmaps overlaid on the original images, with warmer colors indicating higher confidence regions. Results demonstrate rich encoder representations of Anatomy-VLM that effectively capture anatomical knowledge and transfer well to improve segmentation performance.}
  \label{fig:segmentation_compare}
\end{figure*}

\textbf{Evaluation on semantic segmentation.} We evaluate the performance of pretrained visual encoders from different VLMs on the segmentation task by connecting them to an identical U-Net \cite{ronneberger2015u} decoder. We assess two clinically-different tasks: CheXmask for heart segmentation via anatomical labels and SIIM-ACR for pneumonia segmentation via pathological labels. CheXmask provides anatomical segmentation masks derived from the MIMIC-CXR dataset that is seen by all comparison methods. While SIIM-ACR represents a pathological segmentation challenge using the external data that was not seen during training by any of the comparison methods. 

Table \ref{tab:segmentation} presents results under both frozen and fine-tuned encoder settings. For heart segmentation, our method achieves exceptional performance with 0.924 Dice and Mean IoU in the frozen encoder setting, outperforming specialized medical models like BioViL (0.856 Dice, 0.864 Mean IoU) and BioMedCLIP (0.920 Dice, 0.920 Mean IoU). With complete fine-tuning, our approach reaches 0.961 Dice and 0.960 Mean IoU, substantially exceeding all baselines. 

Evaluating pneumonia segmentation is a challenging task that reveals more remarkable differences. In Table \ref{tab:segmentation}, with frozen encoders, Anatomy-VLM achieves 0.153 dice compared to BioViL's 0.085 and BioMedCLIP's 0.051, while CLIP fails at the task given its training on general-domain images. This gap remains large after fine-tuning, where Anatomy-VLM achieves 0.347 Dice compared to the next best baseline of 0.301 from CLIP. We reason that the substantial improvement in pneumonia segmentation can be attributed to Anatomy-VLM's ability to learn fine-grained anatomical information during concept alignment. This alignment enables an understanding of the complex spatial relationships within pneumonia pathology and guides the model to focus on the relevant anatomical regions where pneumonia usually manifests. 

Fig. \ref{fig:segmentation_compare} provides qualitative evidence of segmentation results. In the frozen encoder scenario, CLIP-based models exhibit irregular boundaries (row 1, col. 4) and scattered predictions (row 1, col. 3), demonstrating a limited understanding of cardiac anatomical structure. For the SIIM-ACR dataset, both CLIP and BiomedCLIP fail to identify the relevant regions associated with pneumonia pathology (row 3, cols 3-4), while BioViL incorrectly localizes the area of interest (row 3, col 5). MedKLIP generally captures the primary disease region, but it fails to suppress noises, yielding over-segmentation in unrelated regions (row 3, col. 6 and row 4), showing a lack of disease understanding. These results highlight the importance of fine-grained alignment for the accurate semantic understanding of inherently complex pathological conditions. In the transfer learning scenario, all methods perform well on anatomical segmentation tasks due to the supervisory signal provided during fine-tuning. However, for pathological segmentation, our method demonstrates improved transfer capability in identifying pneumonia-related regions of interest, while baseline methods continue to show inadequate disease understanding. 

The consistent superiority across both anatomical (heart) and pathological (pneumonia) segmentation tasks demonstrates that our concept-based alignment learns more transferable visual representations than standard vision-language pre-training. This finding validates the effectiveness of our training paradigm in capturing localized image features that generalize well to downstream segmentation tasks.


\begin{table}[t]
\centering
\caption{Ablation study of model components.  \checkmark\;indicates a component is present.  Best result per metric is \textbf{bold}.}
\begin{tabular}{ccccc}
\toprule
\multicolumn{3}{c}{Components} & \multicolumn{2}{c}{Performance} \\
\cmidrule(r){1-3} \cmidrule(l){4-5}
Global & Detect & Fine & AUC & mAP \\
\midrule
\checkmark &  &  & 0.82 & --   \\
 & \checkmark &  & --   & 0.71 \\
\checkmark & \checkmark &  & 0.90 & 0.73 \\
\checkmark & \checkmark & \checkmark & \textbf{0.91} & \textbf{0.76} \\
\bottomrule
\end{tabular}
\label{tab:ablation}
\end{table}

\subsection{Analysis}
\textbf{Ablation studies.} We conduct an ablation study that reveals the synergistic effects of our proposed multi-scaling supervision. The encoder with global-only supervision achieves 0.82 AUC. The addition of object detection capabilities significantly enhances performance, boosting both AUC (0.82 to 0.90) and mAP (0.71 to 0.73) for localization tasks, demonstrating that explicit anatomy localizing information substantially improves both classification and localization performance. Most importantly, incorporating fine-grained anatomy-level contrastive learning further improves performance to 0.91 AUC (+10.98\%) and 0.76 mAP (+7.04\%), showing the importance of anatomical level relations in improving diagnostic accuracy in medical imaging. The  performance improvement demonstrates that each component contributes meaningfully to the overall system. In addition, the object detection module provides spatial awareness and localization capabilities that enhance the encoder's global understanding, while the fine-grained anatomical component adds detailed structural knowledge that refines both classification accuracy and localization precision. This complementary relationship suggests that effective multimodal modeling requires the integration of global contextual understanding, spatial localization, and detailed anatomical knowledge, validating our multi-faceted concept alignment in medical imaging.

\textbf{Model performance for fine-grained anatomy to finding matching.} We further delve into the analysis of fine-grained anatomy regions to disease label learning performance. We set up the experiment on anatomy-level matching accuracy between the 29 anatomies and the fine-grained validation labels produced by the label generation algorithm introduced in section \ref{sec:label_generation}. We refer the readers to Table
3 in the supplementary for full results. Anatomy-VLM shows a strong overall performance with an average BMAC of 94.8, AUC of 96.1, and F1 score of 30.3 across 29 anatomical regions. Following expert medical workflows, our framework achieves high-performance anatomical matching through precise bounding box alignment. We recognize that performance variations correlate with the long-tail distribution of anatomical attributes in the training data. For instance, high-frequency regions like ``lung opacity" and ``atelectasis" show superior performance, while under-represented structures such as right atrium (1.3 F1), carina (2.3 F1), and left hemidiaphragm (2.8 F1) exhibit lower F1 scores due to limited training examples. The model's ability to achieve reliable anatomical localization enables more precise anatomy-to-finding matching, since well-localized anatomical regions can better correlate visual findings to their corresponding textual descriptions. Overall, this anatomical precision allows the model to distinguish between subtle pathological variations within anatomical contexts, enhancing diagnostic accuracy with clinically-relevant explanations.
\section{Discussion and Conclusion}
We presented Anatomy-VLM that aligns anatomy and related structure localization  with medical concepts and closely mimics a radiologist's workflow. Anatomy-VLM systematically identifies abnormal findings across different anatomical regions of the X-ray scan,  reasons about how these spatially-distributed observations relate to each other, and arrives at a likely diagnosis. Extensive experimental results show that Anatomy-VLM presents a stepwise decision-making of expert radiologists, delivering higher accuracy, transparent reasoning, and flexible zero-shot explainability. Beyond chest X-ray analysis, the proposed framework’s modular design can be extended to other imaging modalities and clinical tasks. Future investigation will explore applying the strength of Anatomy-VLM to advance downstream real-world clinical tasks. 
{
    \small
    \bibliographystyle{ieeenat_fullname}
    \bibliography{main}
}

\end{document}


\maketitle
\section{Zero-shot Disease Classification Table Expanded Version}
Table \ref{tab:imagenome_noAcc} presents the complete zero-shot performance results for chest image disease classification across all 20 individual disease categories, expanding upon the summarized results shown in Table 1 of the main text. This comprehensive evaluation was conducted on a validation set of 6,223 chest images, testing multiple state-of-the-art vision-language models. The anatomical grouping strategy maps individual diseases to seven higher-level categories based on their primary anatomical involvement and pathophysiological mechanisms. Table \ref{tab:attr_groups} shows the mapping used to group the attributes and the metric calculation.

\begin{table}[h]
  \centering
  \scriptsize
  \setlength{\tabcolsep}{3pt}                
  \renewcommand{\arraystretch}{1.05}         
  \begin{tabular}{@{}>{\raggedright\arraybackslash}p{0.35\linewidth}
                  >{\raggedright\arraybackslash}p{0.60\linewidth}@{}}
    \toprule
    \textbf{Class} & \textbf{Assigned attributes} \\ \midrule
    Lung parenchyma \& air--space disease &
    Consolidation, Lung Opacity, Pulmonary Edema/Hazy Opacity, Lung Lesion, Mass/Nodule (NOS) \\
    Atelectasis \& collapse &
    Atelectasis, Linear/Patchy Atelectasis, Lobar/Segmental Collapse \\
    Pleural space \& pleura &
    Pleural Effusion, Costophrenic Angle Blunting, Pleural/Parenchymal Scarring \\
    Inflation \& airway mechanics &
    Hyperaeration \\
    Diaphragm \& sub‑diaphragmatic &
    Elevated Hemidiaphragm \\
    Cardiomediastinal \& hilar structures &
    Enlarged Cardiac Silhouette, Enlarged Hilum, Vascular Congestion, Tortuous Aorta, Vascular Calcification \\
    Musculoskeletal \& thoracic cage &
    Scoliosis, Spinal Degenerative Changes \\ \bottomrule
  \end{tabular}
  \caption{Chest ImaGenome attributes grouped into higher‑level anatomical classes.}
  \label{tab:attr_groups}
\end{table}

\begin{algorithm}[t]
\caption{Contrastive Text--Label Construction}
\label{alg:contrastive-brief}
\begin{algorithmic}[1]
\REQUIRE Findings $\mathcal{F}=\{f_i\}_{i=1}^{n}$ for $n$ bounding boxes
\ENSURE Sentences $T=\{t_i\}_{i=1}^{n}$, label matrix $L\in\{0,1\}^{n\times n}$

\STATE $L\gets\mathbf 0$; \quad $T\gets\emptyset$

\STATE {\bf Step 1:} \textit{Select one sub‑sentence per box}  
        \FOR{$i=1$ \TO $n$}
            \IF{$f_i$ has sub‑sentences}
                \STATE $t_i\gets$ random sub‑sentence of $f_i$; \ $L[i,i]\gets1$
            \ELSE
                \STATE $t_i\gets\epsilon$
            \ENDIF
        \ENDFOR

\STATE {\bf Step 2:} \textit{Perturb a random 20\% of current sample}  
        \FOR{each $i$ with $L[i,i]=1$}
            \IF{rand$()<0.2$} \STATE
                optionally negate or rephrase $t_i$ and flip $L[i,i]$ if negated
            \ENDIF
        \ENDFOR

\STATE {\bf Step 3:} \textit{Fill empty slots with attribute negatives}  
        \STATE $\mathcal{A}\gets$ attributes found in positive $T$
        \FOR{each $i$ with $t_i=\epsilon$ {\bf and} $\mathcal{A}\neq\emptyset$}
            \STATE $t_i\gets$ positive or negated sentence using random $a\in\mathcal{A}$; \ $L[i,i]\gets0$
        \ENDFOR

\STATE {\bf Step 4:} \textit{Mark duplicates}  
        \FOR{$i<j$} \IF{$t_i=t_j$} \STATE increment $L[i,j]$ and $L[j,i]$ \ENDIF \ENDFOR
\end{algorithmic}
\end{algorithm}

\begin{figure*}[t]              
  \caption{Top 20/45 disease classes ranked to show a long tail distribution for Chest ImaGenome dataset.}
  \centering
  \includegraphics[width=0.9\textwidth]{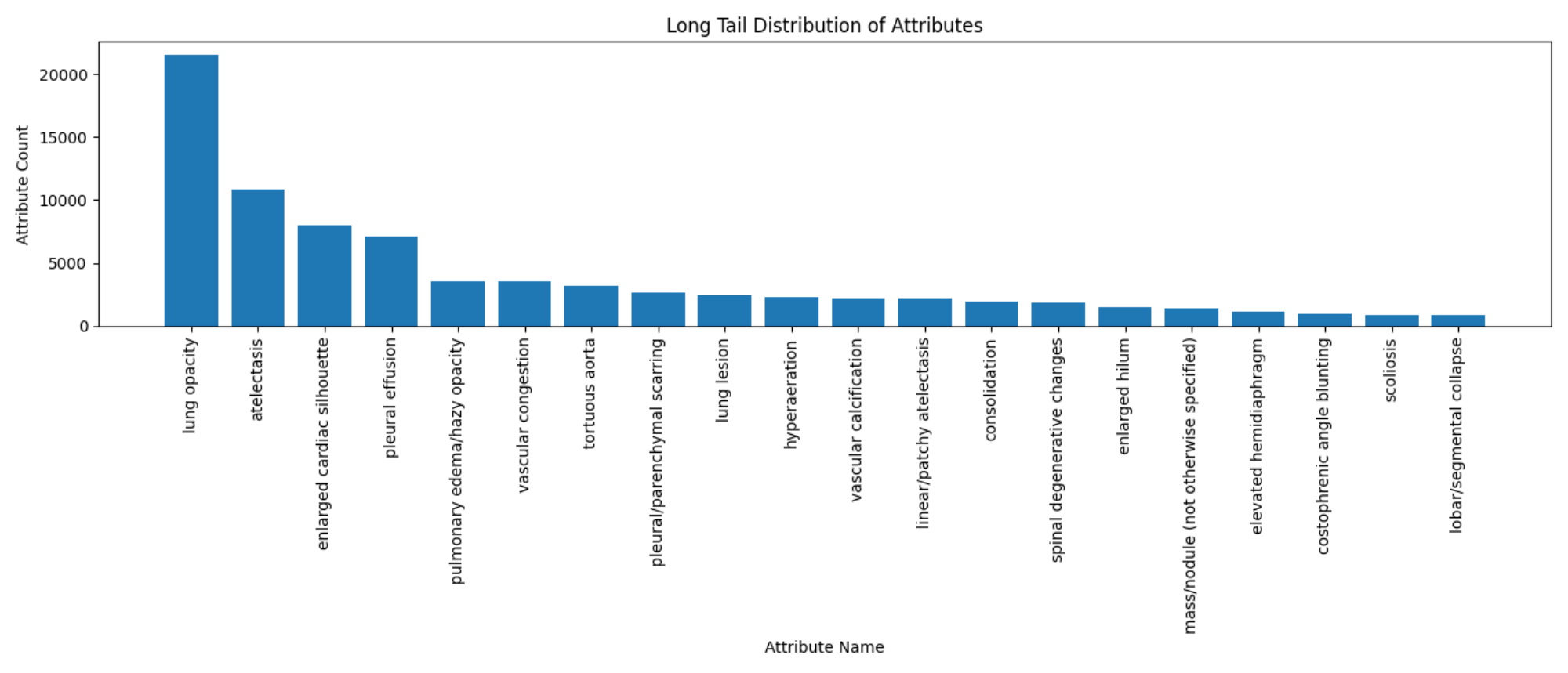}
  \label{fig:long_tail}
\end{figure*}

\begin{figure*}[ht]              
  \caption{25 anatomies to disease class fine-grained learning distribution for Chest ImaGenome dataset.}
  \centering
  \includegraphics[width=0.90\textwidth]{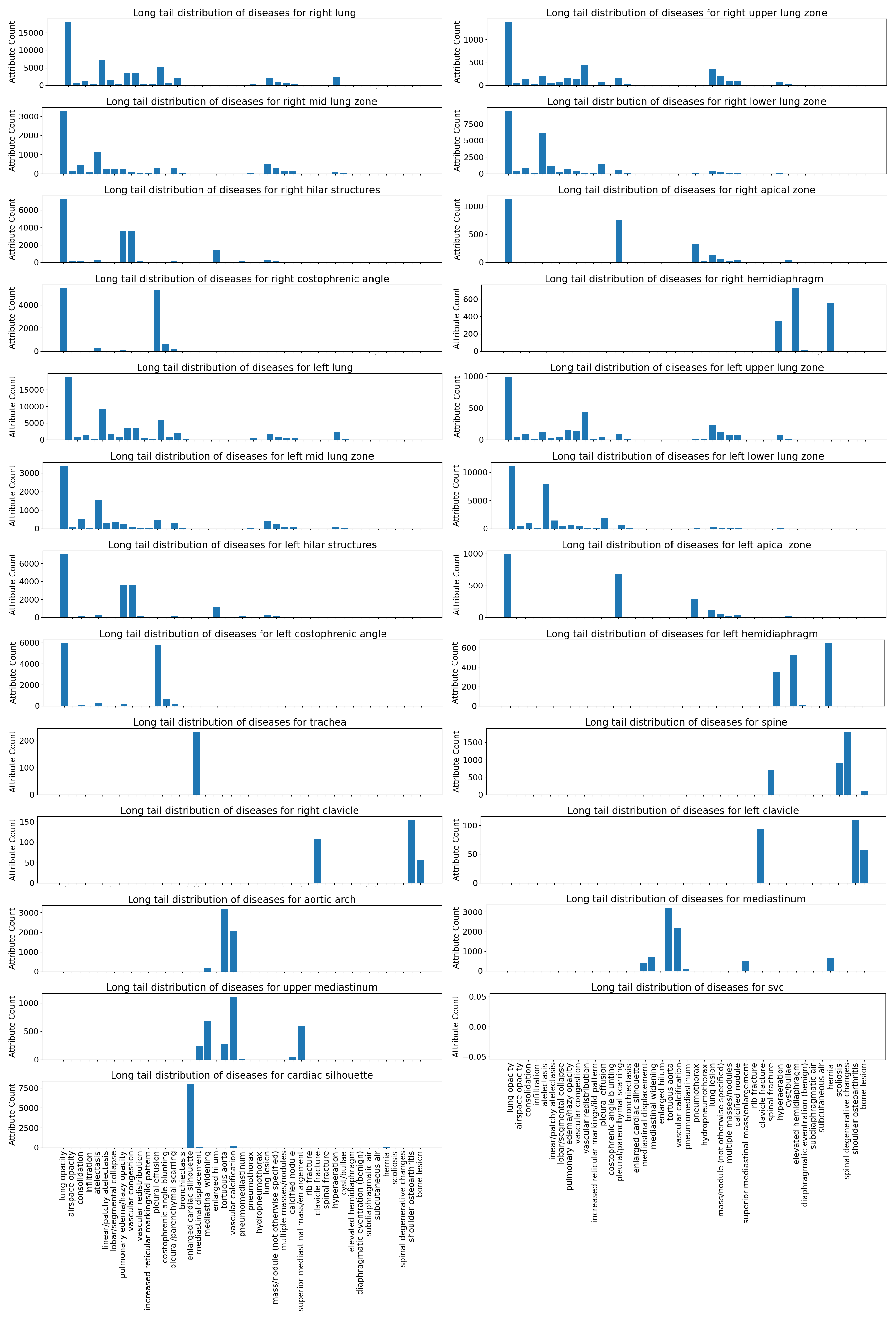}
  \label{fig:fine-grained_data}
\end{figure*}

\begin{table*}[t]
  \centering
  \scriptsize
  \setlength{\tabcolsep}{4pt}
  \renewcommand\theadfont{\normalsize\bfseries}
  \resizebox{\textwidth}{!}{%
  \begin{tabular}{l|ccccccccccccccc|cccccc|ccc}
    \toprule
    \multirow{3}{*}{\textbf{Attribute}} & \multicolumn{15}{c}{\textbf{Zero-shot vision--language models}} & \multicolumn{6}{c}{\textbf{Black-box baselines}} & \multicolumn{3}{c}{\textbf{Concept-aligned (Ours)}} \\[2pt]
    & \multicolumn{3}{c}{CLIP}& \multicolumn{3}{c}{BioMedCLIP}& \multicolumn{3}{c}{BioViL}& \multicolumn{3}{c}{MedKLIP}& \multicolumn{3}{c}{CARZero} & \multicolumn{3}{c}{ResNet}& \multicolumn{3}{c}{ViT}& \multicolumn{3}{c}{Ours} \\[-2pt]
    \cmidrule(lr){2-4}\cmidrule(lr){5-7}\cmidrule(lr){8-10}\cmidrule(lr){11-13}
\cmidrule(lr){14-16}\cmidrule(lr){17-19}\cmidrule(lr){20-22}\cmidrule(lr){23-25}
    & BMAC & AUC & F1 & BMAC & AUC & F1 & BMAC & AUC & F1 & BMAC & AUC & F1 & BMAC & AUC & F1 & BMAC & AUC & F1 & BMAC & AUC & F1 & BMAC & AUC & F1 \\ \midrule
    Atelectasis & 53.0 & 53.0 & 27.2 & 48.1 & 48.1 & 9.0 & 67.3 & 67.3 & 49.8 & 54.1 & 57.0 & 53.7 & 59.6 & 86.4 & 33.1 & 60.1 & 69.9 & 68.1 & 73.1 & 79.7 & 56.6 & 73.4 & 80.8 & 57.4 \\
    Consolidation & 51.5 & 51.5 & 6.8 & 53.7 & 53.7 & 11.9 & 77.8 & 77.8 & 21.6 & 58.5 & 65.7 & 42.4 & 65.2 & 87.3 & 32.6 & 63.0 & 68.0 & 45.9 & 63.8 & 76.8 & 25.8 & 73.4 & 82.5 & 26.3 \\
    Costophrenic Angle Blunting & 50.2 & 50.2 & 3.2 & 49.9 & 49.9 & 0.0 & 64.0 & 64.0 & 6.0 & 52.4 & 66.3 & 8.8 & 66.5 & 83.4 & 15.5 & 59.6 & 63.3 & 12.7 & 63.0 & 74.3 & 10.5 & 59.3 & 78.4 & 13.8 \\
    Elevated Hemidiaphragm & 49.9 & 49.9 & 0.0 & 52.7 & 52.7 & 7.5 & 63.2 & 63.2 & 7.1 & 58.7 & 65.6 & 4.6 & 77.1 & 88.3 & 40.6 & 55.7 & 63.5 & 8.0 & 72.5 & 85.8 & 35.6 & 66.1 & 84.8 & 32.9 \\
    Enlarged Cardiac Silhouette & 50.0 & 50.0 & 0.4 & 52.9 & 52.9 & 14.0 & 75.6 & 75.6 & 52.9 & 51.3 & 66.9 & 4.6 & 65.4 & 91.8 & 46.3 & 54.6 & 63.9 & 9.1 & 76.2 & 87.3 & 59.4 & 77.9 & 89.3 & 61.7 \\
    Enlarged Hilum & 48.1 & 48.1 & 5.6 & 50.1 & 50.1 & 1.0 & 69.2 & 69.2 & 10.7 & 50.0 & 59.8 & 29.7 & 61.6 & 82.9 & 23.2 & 71.1 & 79.8 & 49.3 & 60.8 & 70.2 & 17.4 & 62.7 & 75.2 & 15.5 \\
    Hyperaeration & 50.0 & 50.0 & 10.0 & 50.1 & 50.1 & 0.6 & 58.4 & 58.4 & 12.0 & 50.0 & 43.4 & 5.9 & 78.1 & 90.6 & 45.4 & 56.8 & 56.7 & 8.2 & 71.0 & 87.4 & 39.7 & 74.3 & 89.8 & 46.3 \\
    Linear/Patchy Atelectasis & 49.7 & 49.7 & 3.9 & 48.9 & 48.9 & 2.4 & 57.3 & 57.3 & 11.5 & 41.7 & 37.3 & 7.0 & 63.4 & 75.4 & 20.5 & 57.4 & 64.2 & 16.9 & 60.0 & 66.7 & 15.9 & 66.4 & 72.1 & 18.2 \\
    Lobar/Segmental Collapse & 50.7 & 50.7 & 3.5 & 63.1 & 63.1 & 13.7 & 73.3 & 73.3 & 9.7 & 55.9 & 57.5 & 11.6 & 65.7 & 88.6 & 25.1 & 59.8 & 60.9 & 12.6 & 59.9 & 84.7 & 22.1 & 61.6 & 87.0 & 22.3 \\
    Lung Lesion & 50.9 & 50.9 & 10.7 & 52.0 & 52.0 & 8.7 & 59.1 & 59.1 & 13.8 & 50.0 & 54.7 & 3.9 & 57.0 & 69.0 & 19.7 & 58.3 & 69.0 & 9.9 & 57.0 & 62.6 & 16.4 & 59.0 & 66.6 & 17.4 \\
    Lung Opacity & 49.0 & 49.0 & 64.2 & 54.2 & 54.2 & 17.0 & 74.9 & 74.9 & 69.3 & 50.0 & 59.5 & 10.6 & 55.0 & 87.4 & 18.7 & 51.4 & 49.6 & 10.8 & 76.9 & 85.1 & 77.5 & 77.5 & 86.2 & 78.3 \\
    Mass/Nodule (Not Otherwise Specified) & 50.4 & 50.4 & 5.9 & 50.8 & 50.8 & 5.5 & 60.1 & 60.1 & 7.7 & 50.0 & 50.5 & 5.8 & 64.2 & 78.9 & 27.3 & 52.8 & 50.8 & 6.2 & 55.7 & 63.7 & 12.0 & 59.8 & 70.6 & 12.6 \\
    Pleural Effusion & 50.2 & 50.2 & 27.8 & 52.2 & 52.2 & 8.6 & 81.4 & 81.4 & 59.8 & 50.2 & 65.7 & 28.0 & 62.6 & 93.9 & 40.0 & 71.2 & 78.6 & 46.4 & 82.9 & 90.8 & 66.5 & 80.8 & 91.4 & 67.1 \\
    Pleural/Parenchymal Scarring & 50.0 & 50.0 & 11.4 & 50.3 & 50.3 & 3.7 & 57.6 & 57.6 & 13.9 & 50.7 & 54.3 & 11.5 & 57.6 & 80.2 & 23.1 & 56.2 & 57.9 & 13.5 & 61.3 & 69.6 & 20.3 & 67.4 & 75.2 & 24.2 \\
    Pulmonary Edema/Hazy Opacity & 58.2 & 58.2 & 18.1 & 66.6 & 66.6 & 36.4 & 78.2 & 78.2 & 41.7 & 56.0 & 59.6 & 17.5 & 61.9 & 91.3 & 36.4 & 63.2 & 69.7 & 25.0 & 72.7 & 85.7 & 46.1 & 77.5 & 88.9 & 48.4 \\
    Scoliosis & 50.0 & 50.0 & 4.5 & 50.2 & 50.2 & 1.9 & 48.1 & 48.1 & 4.1 & 49.2 & 46.6 & 4.4 & 74.2 & 83.2 & 27.6 & 58.2 & 63.7 & 11.3 & 62.5 & 80.0 & 31.1 & 60.8 & 78.1 & 22.0 \\
    Spinal Degenerative Changes & 50.0 & 50.0 & 0.0 & 50.0 & 50.0 & 0.0 & 56.1 & 56.1 & 8.7 & 50.0 & 41.3 & 7.5 & 64.7 & 82.8 & 21.7 & 62.4 & 67.5 & 14.1 & 60.6 & 71.9 & 14.1 & 67.0 & 75.5 & 16.4 \\
    Tortuous Aorta & 50.0 & 50.0 & 14.4 & 49.0 & 49.0 & 4.5 & 63.8 & 63.8 & 20.2 & 50.2 & 60.7 & 14.4 & 77.7 & 89.7 & 44.9 & 62.6 & 71.9 & 26.3 & 70.6 & 84.3 & 38.4 & 68.2 & 83.0 & 36.7 \\
    Vascular Calcification & 50.0 & 50.0 & 0.0 & 50.0 & 50.0 & 0.0 & 66.0 & 66.0 & 16.0 & 50.0 & 49.8 & 10.2 & 72.5 & 89.1 & 38.4 & 58.8 & 64.9 & 17.3 & 71.0 & 81.8 & 28.9 & 66.8 & 83.9 & 30.2 \\
    Vascular Congestion & 49.1 & 49.1 & 3.1 & 52.9 & 52.9 & 12.7 & 70.9 & 70.9 & 28.8 & 50.0 & 51.0 & 13.9 & 65.6 & 86.7 & 36.2 & 61.4 & 66.8 & 21.7 & 68.6 & 82.0 & 33.6 & 73.6 & 85.4 & 36.1 \\
    Average & 50.6 & 50.6 & 11.0 & 52.4 & 52.4 & 8.0 & 66.1 & 66.1 & 23.3 & 51.4 & 55.7 & 14.8 & 65.8 & 85.4 & 30.8 & 59.7 & 65.0 & 21.7 & 67.0 & 78.5 & 33.4 & 68.7 & 81.2 & 34.2 \\
    \bottomrule
  \end{tabular}}
  \caption{Chest ImaGenome zero-shot global disease-classification (Full result).}
  \label{tab:imagenome_noAcc}
\end{table*}

\section{Anatomy to Disease Class Distribution}
We analyzed the disease class frequency distribution in the Chest ImaGenome dataset, with results presented in Figure \ref{fig:long_tail}. The disease class labels were counted from the training set, revealing a long-tail distribution that reflects significant class imbalance across different conditions. Additionally, we examined the label distribution for each anatomical region provided by the Chest ImaGenome dataset, as shown in Figure \ref{fig:fine-grained_data}. Note that Cavoatrial Junction, Right Atrium, Carina, and Abdomen are excluded from this analysis as they contain no matching disease findings, similar to the SVC region. The anatomical region analysis confirms that the long-tail distribution persists across different body regions, while also revealing complex disease patterns that provide meaningful training dynamics for Anatomy-VLM development.

\section{Fine-grained Contrastive Learning Label Generation Algorithm}
We provide exact algorithmic procedure in algorithm \ref{alg:contrastive-brief} to detail the systematic procedure for generating fine-grained contrastive learning labels from radiological findings and their corresponding bounding boxes. The algorithm transforms input findings for \(n\) bounding boxes into sentences with corresponding label matrix through a four-step process designed to address key challenges in medical image-text alignment. The procedure begins by selecting representative sentences from available findings for each bounding box, establishing positive correspondences in the label matrix. To create meaningful contrastive pairs, the algorithm then generates negative samples by randomly selecting a subset of positive assignments and applying semantic perturbations such as negation or rephrasing. For bounding boxes lacking specific findings, the algorithm fills empty slots using a predefined attribute set derived from positive training examples, ensuring comprehensive coverage of radiological vocabulary. Finally, the algorithm maintains consistency by identifying duplicate sentences and ensuring they receive identical labels across all positions. This systematic approach captures the nuanced relationship between radiological findings and their textual descriptions while maintaining clinical validity. The algorithm's design reflects the complexity of medical image interpretation, where precise terminology and semantic understanding are crucial for accurate diagnosis. By generating both positive and negative examples with appropriate label assignments, the contrastive learning framework can effectively learn to distinguish between different radiological conditions and their corresponding textual descriptions, ultimately improving the performance of vision-language models in medical imaging applications.

\begin{table*}[h]
  \centering
  \scriptsize
  \caption{Zero-shot region-wise validation (bounding boxes) on Chest ImaGenome dataset}
  \label{tab:zs_bbox_igenome}
  \setlength{\tabcolsep}{10pt}
  \renewcommand{\arraystretch}{1.1}
  \begin{tabular}{lccc}
    \toprule
    \textbf{Region} & \textbf{BMAC} & \textbf{AUC} & \textbf{F1} \\
    \midrule
    Right lung & 98.1 & 98.1 & 66.2 \\
    Right upper lung zone & 84.3 & 90.1 & 6.3 \\
    Right mid lung zone & 95.4 & 96.8 & 13.8 \\
    Right lower lung zone & 94.2 & 95.5 & 15.8 \\
    Right hilar structures & 97.8 & 98.5 & 58.4 \\
    Right apical zone & 97.5 & 97.7 & 24.4 \\
    Right costophrenic angle & 96.6 & 96.7 & 47.2 \\
    Right hemidiaphragm & 92.4 & 93.8 & 5.5 \\
    Left lung & 98.0 & 98.2 & 65.7 \\
    Left upper lung zone & 84.9 & 90.6 & 4.5 \\
    Left mid lung zone & 95.3 & 96.6 & 16.0 \\
    Left lower lung zone & 93.9 & 95.4 & 17.7 \\
    Left hilar structures & 97.9 & 98.4 & 58.5 \\
    Left apical zone & 97.4 & 98.0 & 24.8 \\
    Left costophrenic angle & 96.7 & 96.8 & 47.7 \\
    Left hemidiaphragm & 90.8 & 93.1 & 2.8 \\
    Trachea & 94.7 & 95.9 & 5.2 \\
    Spine & 99.1 & 99.6 & 96.4 \\
    Right clavicle & 98.2 & 99.3 & 32.9 \\
    Left clavicle & 98.1 & 99.3 & 33.2 \\
    Aortic arch & 91.7 & 94.7 & 6.7 \\
    Mediastinum & 98.2 & 98.5 & 64.8 \\
    Upper mediastinum & 96.9 & 97.9 & 48.8 \\
    SVC & 92.9 & 94.4 & 2.6 \\
    Cardiac silhouette & 99.7 & 99.7 & 98.2 \\
    Cavoatrial junction & 88.1 & 92.4 & 0.4 \\
    Right atrium & 90.6 & 91.6 & 1.3 \\
    Carina & 94.5 & 96.3 & 2.3 \\
    Abdomen & 94.1 & 94.2 & 11.2 \\
    \midrule
    \textbf{Average} & \textbf{94.8} & \textbf{96.1} & \textbf{30.3} \\
    \bottomrule
  \end{tabular}
\end{table*}    